\documentclass[11pt,a4paper]{article}
\usepackage[hyperref]{acl2020}
\usepackage{times}
\usepackage{latexsym}

\usepackage{microtype}
\aclfinalcopy 
\usepackage{multirow}

\usepackage{graphicx} 

\usepackage{caption}
\usepackage[normalem]{ulem}
\usepackage[utf8]{inputenc}
\usepackage{dingbat}
\usepackage{wrapfig}
\usepackage{amssymb}
\usepackage{amsfonts}
\usepackage{amsmath}
\usepackage{amssymb}
\usepackage{amsthm}
\usepackage{mathrsfs}
\usepackage{adjustbox}
\usepackage{bbold}
\usepackage[font=small]{caption}
\usepackage[linesnumbered,ruled]{algorithm2e}
\usepackage{url}

\DeclareMathOperator*{\argmin}{argmin}

\usepackage{enumitem}
\usepackage{wrapfig,lipsum,booktabs}

\usepackage[linesnumbered,ruled]{algorithm2e}
\usepackage{adjustbox}
\usepackage{mathtools}

\newcommand{\multidds}{MultiDDS}

\title{Balancing Training for \\ Multilingual Neural Machine Translation}

\author{Xinyi Wang \qquad Yulia Tsvetkov \qquad Graham Neubig \\
 Language Technology Institute, Carnegie Mellon University, Pittsburgh, PA 15213 \\
 {\texttt{\{xinyiw1,ytsvetko,gneubig\}@cs.cmu.edu}} }

\setlist[description]{font=\normalfont\itshape\space}

\begin{document}
\maketitle

\begin{abstract}
    When training multilingual machine translation (MT) models that can translate to/from multiple languages, we are faced with imbalanced training sets: some languages have much more training data than others.
    Standard practice is to up-sample less resourced languages to increase representation, and the degree of up-sampling has a large effect on the overall performance.
    In this paper, we propose a method that instead automatically learns how to weight training data through a data scorer that is optimized to maximize performance on all test languages.
    Experiments on two sets of languages under both one-to-many and many-to-one MT settings show our method not only consistently outperforms heuristic baselines in terms of average performance, but also offers flexible control over the performance of which languages are optimized.%
    \footnote{The code is available at \url{https://github.com/cindyxinyiwang/multiDDS/}.}   
\end{abstract}

\section{Introduction}

Multilingual models are trained to process different languages in a single model, and have been applied to a wide variety of NLP tasks such as text classification \citep{klementiev2012inducing,chen2018adversarial}, syntactic analysis \citep{plank-etal-2016-multilingual,ammar2016many}, named-entity recognition~\citep{xie-etal-2018-neural,wu-dredze-2019-beto}, and machine translation~(MT)~\citep{dong-etal-2015-multi,johnson16multilingual}.
These models have two particularly concrete advantages over their monolingual counterparts.
First, deploying a single multilingual model is much more resource efficient than deploying one model for each language under consideration~\citep{massive_wild,massive}.
Second, multilingual training makes it possible to transfer knowledge from high-resource languages~(HRLs) to improve performance on low-resource languages~(LRLs)~\cite{multi_nmt_adapt,multi_nmt_bpe_share,rapid_adapt_nmt,tcs,massive}.  

A common problem with multilingual training is that the data from different languages are both heterogeneous (different languages may exhibit very different properties) and imbalanced (there may be wildly varying amounts of training data for each language).
Thus, while LRLs will often benefit from transfer from other languages, for languages where sufficient monolingual data exists, performance will often \emph{decrease} due to interference from the heterogeneous nature of the data.
This is especially the case for modestly-sized models that are conducive to efficient deployment \citep{massive_wild,xlmr}.

To balance the performance on different languages, the standard practice is to heuristically adjust the distribution of data used in training, specifically by over-sampling the training data from LRLs \citep{johnson16multilingual,rapid_adapt_nmt,massive_wild,xlmr}.
For example, \citet{massive_wild} sample training data from different languages based on the dataset size scaled by a heuristically tuned temperature term.
However, such heuristics are far from perfect.
First, \citet{massive_wild} find that the exact value of this temperature term significantly affects results, and we further show in experiments that the ideal temperature varies significantly from one experimental setting to another.
Second, this heuristic ignores factors other than data size that affect the interaction between different languages, despite the fact that language similarity has been empirically proven important in examinations of cross-lingual transfer learning~\citep{tcs,choose_transfer}.

In this paper, we ask the question: ``is it possible to \emph{learn} an optimal strategy to automatically balance the usage of data in multilingual model training?''
To this effect, we propose a method that learns a language scorer that can be used throughout training to improve the model performance on \textit{all} languages.
Our method is based on the recently proposed approach of Differentiable Data Selection~\citep[DDS]{dds}, a general machine learning method for optimizing the weighting of different training examples to improve a pre-determined objective.
In this work, we take this objective to be the average loss from different languages, and directly optimize the weights of training data from each language to maximize this objective on a multilingual development set.
This formulation has no heuristic temperatures, and enables the language scorer to consider the interaction between languages.

Based on this formulation, we propose an algorithm that improves the ability of DDS to optimize \emph{multiple} model objectives, which we name \multidds.
This is particularly useful in the case where we want to optimize performance on multiple languages simultaneously.
Specifically, \multidds~(1) has a more flexible scorer parameterization, (2) is memory efficient when training on multiple languages, and (3) stabilizes the reward signal so that it improves all objectives simultaneously instead of being overwhelmed by a single objective. 

While the proposed methods are model-agnostic and thus potentially applicable to a wide variety of tasks, we specifically test them on the problem of training multilingual NMT systems that can translate many languages in a single model.
We perform experiments on two sets of languages (one with more similarity between the languages, one with less) and two translation directions (one-to-many and many-to-one where the ``one'' is English).
Results show that  \multidds~consistently outperforms various baselines in all settings.
Moreover, we demonstrate \multidds~provides a flexible framework that allows the user to define a variety of optimization objectives for multilingual models. 

\section{\label{sec:method}Multilingual Training Preliminaries}

\paragraph{Monolingual Training Objective}
A standard NMT model is trained to translate from a single source language $S$ to a target language $T$.
The parameters of the model are generally trained by preparing a training dataset $D_{\text{train}}$, and defining the empirical distribution of sentence pairs $\langle x, y \rangle$ sampled from $D_{\text{train}}$ as $P$.
We then minimize the empirical risk $J(\theta, P)$, which is the expected value of the loss function $\ell(x, y; \theta)$ over this distribution: 
\begin{align}
\label{eqn:obj_mono_nmt}
\begin{split}
    & \theta^* = \argmin_\theta J(\theta, D_{\text{train}}) \\
    & \text{where}~~~ J(\theta, D_{\text{train}}) = \mathbb{E}_{x, y \sim P(X, Y)}[\ell(x, y;\theta)]
\end{split}
\end{align}

\paragraph{Multilingual Training Formulation}
A multilingual NMT model can translate $n$ pairs of languages $\{S^1\text{-}T^1, S^2\text{-}T^2, ..., S^n\text{-}T^n\}$, from any source language $S^i$. to its corresponding target $T^i$. To train such a multilingual model, we have access to $n$ sets of training data $D_\text{train} = D_\text{train}^1, D_\text{train}^2, \ldots, D_\text{train}^n$, where $D_\text{train}^i$ is training data for language pair $S^i\text{-}T^i$.
From these datasets, we can define $P^i$, the distribution of sentences from $S^i$-$T^i$, and consequently also define a risk $J(\theta, P^i)$ for each language following the monolingual objective in \autoref{eqn:obj_mono_nmt}.

However, the question now becomes: ``how do we define an overall training objective given these multiple separate datasets?''
Several different methods to do so have been proposed in the past.
To discuss all of these different methods in a unified framework, we further define a distribution $P_D$ over the $n$ sets of training data, and define our overall multilingual training objective as
\begin{align}
    \label{eqn:obj_multilin}
    J_{\text{mult}}(\theta, P_D, D_{\text{train}}) =  \mathbb{E}_{i \sim P_D(i; \psi)} \left[ J(\theta, D_{\text{train}}^i) \right]. 
\end{align}
In practice, this overall objective can be approximated by selecting a language according to $\tilde{i} \sim P_D(i)$, then calculating gradients with respect to $\theta$ on a batch of data from $D_\text{train}^{\tilde{i}}$.

\paragraph{Evaluation Methods}
 Another important question is how to evaluate the performance of such multilingual models.
 During training, it is common to use a separate development set for each language $D_\text{dev} = D_\text{dev}^1, D_\text{dev}^2, ..., D_\text{dev}^n$ to select the best model.
 Given that the objective of multilingual training is generally to optimize the performance on all languages simultaneously~\citep{massive_wild,xlmr}, we can formalize this objective as minimizing the average of dev risks\footnote{In reality, it is common to have the loss $\ell$ be a likelihood-based objective, but finally measure another metric such as BLEU score at test time, but for simplicity we will assume that these two metrics are correlated.}:
\begin{equation}
\label{eqn:dev_agg}
J_{\text{dev}}(\theta, D_{\text{dev}}) = \frac{1}{n} \sum_{i=1}^n J(\theta, D_{\text{dev}}^i).
\end{equation}



\paragraph{Relation to Heuristic Strategies}
This formulation generalizes a variety of existing techniques that define $P_D(i)$ using a heuristic strategy, and keep it fixed throughout training.

\begin{description}[leftmargin=5pt]
\item[Uniform:] The simplest strategy sets $P_D(i)$ to a uniform distribution, sampling minibatches from each language with equal frequency \cite{johnson16multilingual}.
\item[Proportional:] It is also common to sample data in portions equivalent to the size of the corresponding corpora in each language \cite{johnson16multilingual,rapid_adapt_nmt}. 
\item[Temperature-based:] Finally, because both of the strategies above are extreme (proportional under-weighting LRLs, and uniform causing overfitting by re-sampling sentences from limited-size LRL datasets), it is common to sample according to data size exponentiated by a temperature term $\tau$ \cite{massive_wild,xlmr}:
\begin{align}
    \label{eqn:temp_data}
   P_D(i) = \frac{q_i^{1/\tau}}{\sum_{k=1}^n q_k^{1/\tau}}~\text{where}~q_i = \frac{|D_\text{train}^i|}{\sum_{k=1}^n|D_\text{train}^k|}.
\end{align}
When $\tau=1$ or $\tau=\infty$ this is equivalent to proportional or uniform sampling respectively, and when a number in the middle is chosen it becomes possible to balance between the two strategies.
\end{description}

As noted in  the introduction, these heuristic strategies have several drawbacks regarding sensitivity to the $\tau$ hyperparameter, and lack of consideration of similarity between the languages.
In the following sections we will propose methods to resolve these issues.


\section{Differentiable Data Selection}
Now we turn to the question: is there a better way to optimize $P_D(i)$ so that we can achieve our final objective of performing well on a representative development set over all languages, i.e.~minimizing $J_{\text{dev}}(\theta, D_{\text{dev}})$.
In order to do so, we turn to a recently proposed method of Differentiable Data Selection~\citep[DDS]{dds}, a general purpose machine learning method that allows for weighting of training data to improve performance on a separate set of held-out data.

Specifically, DDS uses a technique called \emph{bi-level optimization} \cite{bilevel_optim}, that learns a second set of parameters $\psi$ that modify the training objective that we use to learn $\theta$, so as to maximize the final objective $J_{\text{dev}}(\theta, D_{\text{dev}})$.
Specifically, it proposes to learn a data scorer $P(x, y; \psi)$, parameterized by $\psi$, such that training using data sampled from the scorer optimizes the model performance on the dev set.
To take the example of learning an NMT system to translate a \emph{single} language pair $i$ using DDS, the general objective in \autoref{eqn:obj_mono_nmt} could be rewritten as
\begin{align}
    \begin{split}
   & \psi^* = \argmin_\psi  J(\theta^*(\psi), D_\text{dev}^i)  ~~\text{where}~~ \\
    &\theta^*(\psi) = \argmin_\theta \mathbb{E}_{x,y \sim P(x, y; \psi)} \left[ \ell(x, y;\theta) \right]. 
\end{split}
\end{align}



DDS optimizes $\theta$ and $\psi$ iteratively throughout the training process. Given a fixed $\psi$, the update rule for $\theta$ is simply
\begin{align*}
    \theta_t \leftarrow \theta_{t-1} - \nabla_\theta 
    \mathbb{E}_{x,y \sim P(x,y; \psi)} \left[ \ell(x, y; \theta) \right]
\end{align*}

To update the data scorer, DDS uses reinforcement learning with a reward function that approximates the effect of the training data on the model's dev performance
\begin{align}
\begin{split}
    R(x,y;\theta_t) &\approx  \nabla J(\theta_t, D_\text{dev}^i)^\top \cdot \nabla_\theta \ell(x, y; \theta_{t-1})  \\
   & \approx \text{cos} \left( \nabla J(\theta_t, D_\text{dev}^i), \nabla_\theta \ell(x, y; \theta_{t-1}) \right)
\end{split}
\end{align}
where $\text{cos}(\cdot)$ is the cosine similarity of two vectors. This reward can be derived by directly differentiating $J(\theta(\psi), D_\text{dev}^i)$ with respect to $\psi$, but intuitively, it indicates that the data scorer should be updated to up-weigh the data points that have similar gradient with the dev data.
According to the REINFORCE algorithm~\cite{reinforce}, the update rule for the data scorer then becomes 
\begin{align}
    \psi_{t+1} \leftarrow \psi_t + R(x, y;\theta_t) \cdot \nabla_\psi \text{log} P(x, y; \psi) 
\end{align}

\section{DDS for Multilingual Training}
\begin{figure}
    \centering
    \includegraphics[width=0.9\columnwidth]{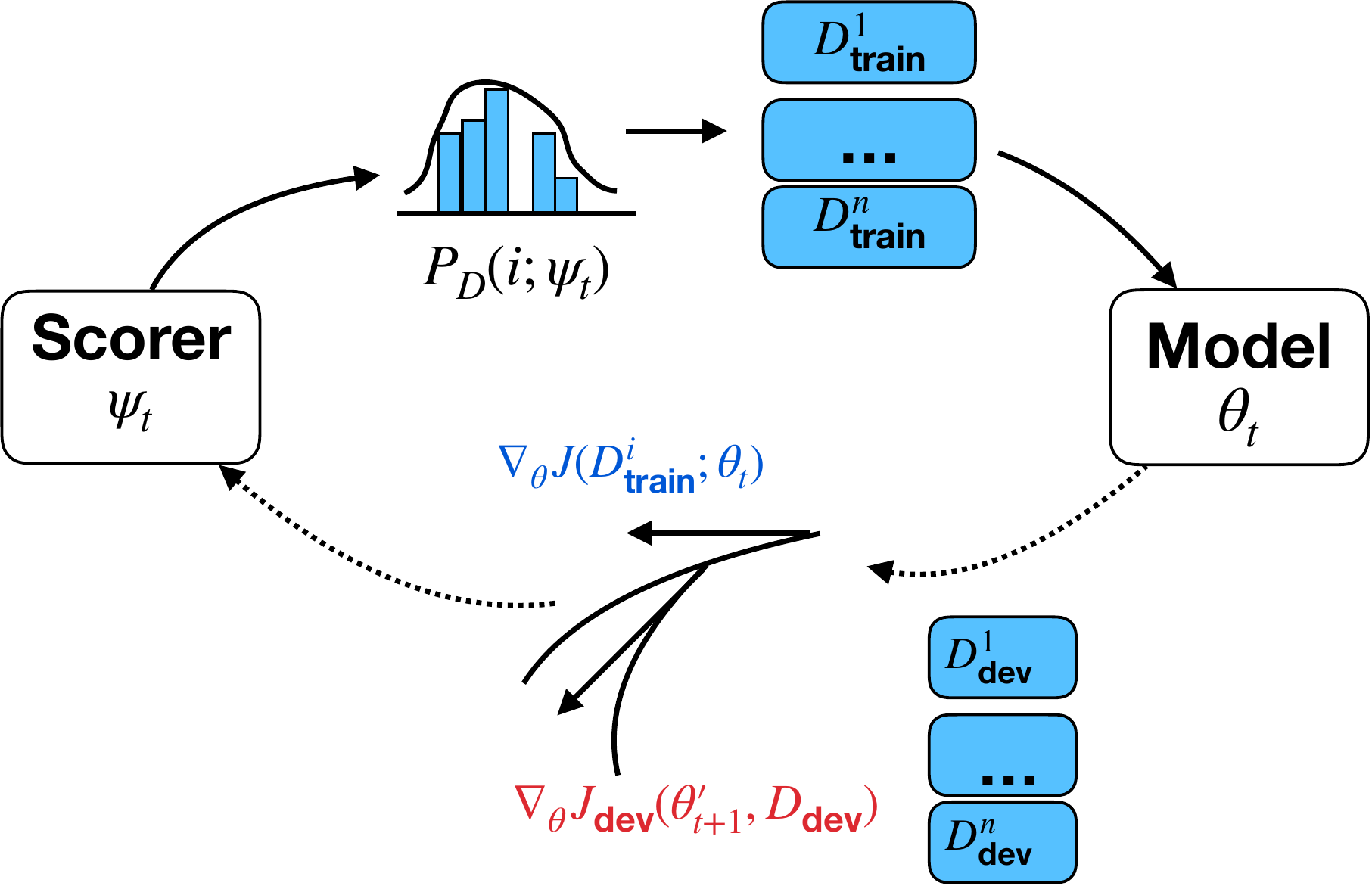}
    \caption{An illustration of the \multidds~algorithm. Solid lines represent updates for $\theta$, and dashed lines represent updates for $\psi$. The scorer defines the distribution over $n$ training languages, from which training data is sampled to train the model. The scorer is updated to favor the datasets with similar gradients as the gradient of the aggregated dev sets.}
    \label{fig:multidds}
\end{figure}

In this section, we use the previously described DDS method to derive a new framework that, instead of relying on fixed heuristics, adaptively optimizes usage of multilingual data for the best model performance on \textit{multiple} languages.
We illustrate the overall workflow in \autoref{fig:multidds}.

First, we note two desiderata for our multilingual training method: 1) \textbf{generality}: the method should be flexible enough so that it can be utilized universally for different multilingual tasks and settings~(such as different translation directions for NMT). 2) \textbf{scalablity}: the method should be stable and efficient if one wishes to scale up the number of languages that a multilingual model supports.
Based on these two properties, we introduce \multidds, an extension of the DDS method tailored for multilingual training.

\paragraph{Method} \multidds~directly parameterizes the standard dataset sampling distribution for multilingual training with $\psi$:
\begin{equation}
\label{eqn:scorer_param}
    P_D(i; \psi) = e^{\psi_i} / \textstyle{\sum_{k=1}^n} e^{\psi_{k}}
\end{equation}
and optimizes $\psi$ to minimize the dev loss.
Notably, unlike standard DDS we make the design decision to weight training datasets rather than score each training example $\langle x, y \rangle$ directly, as it is more efficient and also likely easier to learn.

We can thus rewrite the objective in \autoref{eqn:obj_multilin} to incorporate both $\psi$ and $\theta$ as:

\begin{align}
\label{eqn:obj_multilin_psi_theta}
\begin{split}
   & \psi^* = \argmin_\psi J_\text{dev}(\theta^*(\psi), D_\text{dev})  ~~\text{where}~~ \\
    &\theta^* = \argmin_\theta \mathbb{E}_{i \sim P_D(i; \psi)} \left[ J(\theta, D_\text{train}^i) \right] 
\end{split}
\end{align}
In other words, while the general DDS framework evaluates the model performance on a single dev set and optimizes the weighting of each training example, our multilingual training objective evaluates the performance over an aggregation of $n$ dev sets and optimizes the weighting of $n$ training sets.


The reward signal for updating $\psi_t$ is 
 \begin{align}
    \label{eqn:reward}
    \small
    \begin{split}
     & R(i; \theta_t) \approx \text{cos} \left( \nabla \left( J_\text{dev}(\theta_t, D_\text{dev}) \right), \nabla_\theta J(\theta_{t-1}, D_\text{train}^i)  \right) \\
     & = \text{cos} \left( \nabla  \left(\frac{1}{n} \sum_{k=1}^n J(\theta_t, D_\text{dev}^k) \right), \nabla_\theta J(\theta_{t-1}, D_\text{train}^i)  \right),
    \end{split}
 \end{align}
where $J_\text{dev}(\cdot)$ defines the combination of $n$ dev sets, and we simply plug in its definition from \autoref{eqn:dev_agg}. Intuitively, \autoref{eqn:reward} implies that we should favor the training language $i$ if its gradient aligns with the gradient of the aggregated dev risk of all languages. 

\paragraph{Implementing the Scorer Update}
The pseudo-code for the training algorithm using \multidds~can be found in  \autoref{alg:nmt_dds}.
Notably, we do not update the data scorer $\psi$ on every training step, because it is too computationally expensive for NMT training~\citep{dds}. Instead, after training the multilingual model $\theta$ for a certain number of steps, we update the scorer for all languages. This implementation is not only efficient, but also allows us to re-estimate more frequently the effect of languages that have low probability of being sampled.  

In order to do so, it is necessary to calculate the effect of each training language on the current model, namely $R(i; \theta_t)$. We estimate this value by sampling a batch of data from each $D_\text{train}^i$ to get the training gradient for $\theta_t$, and use this to calculate the reward for this language.
This process is detailed in \autoref{alg:stepahead} of the \autoref{alg:nmt_dds}.

Unlike the algorithm in DDS which requires storing $n$ model gradients,%
\footnote{The NMT algorithm in \citep{dds} estimates the reward by storing the moving average of $n$ training gradients, which is not memory efficient~(See Line. 7 of Alg. 2 in \citep{dds}). In the preliminary experiments, our approximation performs as well as the moving average approximation~(see App. \ref{app:stepahead}). Thus, we use our approximation method as the component for \multidds~for the rest of the experiments.}
this approximation does not require extra memory even if $n$ is large, which is important given recent efforts to scale multilingual training to 100+ ~\citep{massive_wild,massive} or even 1000+ languages~\cite{ostling-tiedemann-2017-continuous,malaviya-etal-2017-learning}.

\begin{algorithm}[t!]
\SetAlgoLined
\DontPrintSemicolon
\SetKwInOut{Input}{Input}
\SetKwInOut{Output}{Output}
\SetCommentSty{itshape}
\SetKwComment{Comment}{$\triangleright$ }{}
\Input{$\mathcal{D}_{\text{train}}$; M: amount of data to train the multilingual model before updating $\psi$; 
}
\Output{The converged multilingual model $\theta^*$}
   
  \Comment{Initialize $P_D(i, \psi)$ to be proportional to dataset size}
  $P_D(i, \psi) \leftarrow \frac{|D_{\text{train}}^i|}{\sum_{j=1}^n |D_{\text{train}}^j|}$
  
  \While{not converged}{
    
    \Comment{Load training data with $\psi$}
    $X, Y \leftarrow \emptyset$
    
    \While{$|X, Y| < M$}{
    
        $\tilde{i} \sim P_D(i, \psi_t)$
        
        $(x, y) \sim D_{\text{train}}^{\tilde{i}}$
        
        $X, Y \leftarrow X, Y \cup x, y$
        
    }
    
    \Comment{Train the NMT model for multiple steps}
    \For{$x, y$ in $X, Y$}{
      $\theta \leftarrow \text{GradientUpdate}\left( \theta, \nabla_{\theta} \ell(x, y; \theta) \right)$
        
    }
    \Comment{Estimate the effect of each language $R(i; \theta)$}
    \label{alg:stepahead}
    \For{i from 1 to n}{
        $x', y' \sim D_{\text{train}}^i$
        
        $g_{\text{train}} \leftarrow  \nabla_{\theta} \ell(x', y'; \theta)$
        
        $\theta' \leftarrow \text{GradientUpdate}(\theta,  g_{\text{train}})$
        
        $g_{\text{dev}} \leftarrow 0$
        
        \For{j from 1 to n}{ 
            $x_d, y_d \sim D_{\text{dev}}^j$
            
            $g_{\text{dev}} \leftarrow g_{\text{dev}} +  \nabla_{\theta'} \ell(x_d, y_d; \theta')$
        }
        
        $R(i; \theta) \leftarrow \text{cos}(g_{\text{dev}}, g_{\text{train}})$
    }
    
    \Comment{Optimize $\psi$}
    $d_\psi \leftarrow \sum_{i=1}^n R(i; \theta) \cdot \nabla_{\psi} \text{log}\left( P_D\left( i;\psi \right) \right)$
     
    $\psi \leftarrow \text{GradientUpdate}(\psi, d_{\psi})$ 
  }
  \caption{\label{alg:nmt_dds} Training with \multidds}
\end{algorithm}

 \section{\label{sec:stable_reward}Stabilized Multi-objective Training}
In our initial attempts to scale DDS to highly multi-lingual training, we found that one challenge was that the reward for updating the scorer became unstable.
This is because the gradient of a multilingual dev set is less consistent and of higher variance than that of a monolingual dev set, which influences the fidelity of the data scorer reward.
\footnote{Suppose the dev set gradient of language $k$ has variance of $\text{var}(g_\text{dev}^k) = \sigma$, and that the dev gradients of each language $\{g_\text{dev}^1, ..., g_\text{dev}^n\}$ are independent. Then the sum of the gradients from the $n$ languages has a variance of $\text{var}(\sum_{k=1}^ng_\text{dev}^k) = n\sigma$.}
 
Thus, instead of using the gradient alignment between the training data and the aggregated loss of $n$ dev sets as the reward, we propose a second approach to first calculate the gradient alignment reward between the data and each of the $n$ dev sets, then take the average of these as the final reward.
This can be expressed mathematically as follows:
 \begin{align}
    \label{eqn:stable_reward}
    \begin{split}
   & R'(i; \theta_t) \approx \\
  & \text{cos}\left( \nabla_\theta \left(\frac{1}{n} \sum_{k=1}^n J(\theta_t, D_\text{dev}^k) \right), \nabla_\theta J(\theta_{t-1}, D_\text{train}^i) \right) \\
  & \approx \frac{1}{n} \sum_{k=1}^n \text{cos}\left(\nabla_\theta J(\theta_t, D_\text{dev}^k), \nabla_\theta J(\theta_{t-1}, D_\text{train}^i)\right)
    \end{split}
 \end{align}
 
 To implement this, we can simply replace the standard reward calculation at Line \ref{alg:stepahead} of \autoref{alg:nmt_dds} to use the stable reward.
 We name this setting \multidds-S.
 In \autoref{sec:variance} we show that this method has less variance than the reward in \autoref{eqn:reward}.

\section{Experimental Evaluation}
\subsection{Data and Settings} We use the 58-languages-to-English parallel data from \citet{ted_pretrain_emb}. A multilingual NMT model is trained for each of the two sets of language pairs with different level of language diversity:
\begin{description}
\itemsep-1mm
\item[Related:] 4 LRLs (Azerbaijani: \texttt{aze}, Belarusian: \texttt{bel}, Glacian: \texttt{glg}, Slovak: \texttt{slk}) and a related HRL for each LRL (Turkish: \texttt{tur}, Russian: \texttt{rus}, Portuguese: \texttt{por}, Czech: \texttt{ces})
\item[Diverse:] 8 languages with varying amounts of data, picked without consideration for relatedness (Bosnian: \texttt{bos}, Marathi: \texttt{mar}, Hindi: \texttt{hin}, Macedonian: \texttt{mkd}, Greek: \texttt{ell}, Bulgarian: \texttt{bul}, French: \texttt{fra}, Korean: \texttt{kor})
\end{description}
Statistics of the datasets are in \autoref{app:stats}.

For each set of languages, we test two varieties of translation: 1) many-to-one~(M2O): translating 8 languages to English; 2) one-to-many~(O2M): translating English into 8 different languages. A target language tag is added to the source sentences for the O2M setting~\citep{johnson16multilingual}. 

\subsection{Experiment Setup} All translation models use standard transformer models~\cite{transformer} as implemented in fairseq~\citep{fairseq} with 6 layers and 4 attention heads. All models are trained for 40 epochs. We preprocess the data using sentencpiece~\cite{sentencepiece} with a vocabulary size of $8K$ for each language. The complete set of hyperparameters can be found in \autoref{app:hparams}. The model performance is evaluated with BLEU score~\cite{bleu}, using sacreBLEU~\cite{sacrebleu}.

\paragraph{Baselines} We compare with the three standard heuristic methods explained in \autoref{sec:method}: 
1) Uniform ($\tau=\infty$): datasets are sampled uniformly, so that LRLs are over-sampled to match the size of the HRLs; 2) Temperature: scales the proportional distribution by $\tau=5$ (following \citet{massive_wild}) to slightly over-sample the LRLs; 3) Proportional ($\tau=1$): datasets are sampled proportional to their size, so that there is no over-sampling of the LRLs.  

\paragraph{Ours} we run \multidds~with either the standard reward~(\multidds), or the stabilized reward proposed in \autoref{eqn:stable_reward}~(\multidds-S). The scorer for \multidds~simply maps the ID of each dataset to its corresponding probability (See \autoref{eqn:scorer_param}. The scorer has N parameters for a dataset with N languages.)

\subsection{Main Results}
We first show the average BLEU score over all languages for each translation setting in \autoref{tab:ave_all}.
First, comparing the baselines, we can see that there is no consistently strong strategy for setting the sampling ratio, with proportional sampling being best in the M2O setting, but worst in the O2M setting.
Next, we can see that \multidds~outperforms the best baseline in three of the four settings and is comparable to proportional sampling in the last M2O-Diverse setting.
With the stabilized reward, \multidds-S consistently delivers better overall performance than the best baseline, and outperforms \multidds~in three settings.
From these results, we can conclude that \multidds-S provides a stable strategy to train multilingual systems over a variety of settings.

\begin{table}[ht]
    \centering
    \resizebox{0.5\textwidth}{!}{
    \begin{tabular}{l|l|rr|rr}
    \toprule
       & \multirow{2}{*}{\textbf{Method}}   & \multicolumn{2}{c|}{\textbf{M2O}} & \multicolumn{2}{c}{\textbf{O2M}} \\
       &   & \textbf{Related} & \textbf{Diverse} & \textbf{Related} & \textbf{Diverse} \\
    \midrule
    \multirow{3}{*}{\rotatebox{90}{Baseline}}   &  Uni. ($\tau$=$\infty$) & 22.63 & 24.81 & 15.54 & 16.86 \\ 
       &  Temp. ($\tau$=5)  & 24.00 & 26.01 & 16.61 & 17.94 \\ 
       &  Prop. ($\tau$=1)  & 24.88 & 26.68 & 15.49 & 16.79 \\ 
    \midrule
    \multirow{2}{*}{\rotatebox{90}{Ours}}   &  \multidds  & 25.26 & 26.65 & 17.17 & \textbf{18.40} \\ 
       &  \multidds-S  & \textbf{25.52} & \textbf{27.00} & \textbf{17.32} & 18.24 \\ 
    \bottomrule
    \end{tabular}
    }
    \caption{Average BLEU for the baselines and our methods. Bold indicates the highest value.}
    \label{tab:ave_all}
\end{table}

Next, we look closer at the BLEU score of each language pair for \multidds-S and the best baseline. The results for all translation settings are in \autoref{tab:detail_bleu}. In general, \multidds-S outperforms the baseline on more languages. In the best case, for the O2M-Related setting, \multidds-S brings significant gains for five of the eight languages, without hurting the remaining three. The gains for the Related group are larger than for the Diverse group, likely because \multidds~can take better advantage of language similarities than the baseline methods.

It is worth noting that \multidds~does not impose large training overhead. For example, for our M2O system, the standard method needs around 19 hours and MultiDDS needs around 20 hours for convergence. The change in training time is not siginificant because \multidds~only optimizes a simple distribution over the training datasets.  

\begin{table*}[htbp]
    \centering
    \small
    \begin{tabular}{l|l|l|cccccccc}
    \toprule
     &  \textbf{Method}  & \textbf{Avg.} & \textbf{aze} & \textbf{bel} & \textbf{glg} & \textbf{slk} & \textbf{tur} & \textbf{rus} & \textbf{por} & \textbf{ces}  \\
    \midrule
    \multirow{2}{*}{M2O} & Prop.   & 24.88 & 11.20 & 17.17 & 27.51 & 28.85 & $\textbf{23.09}^*$ & \textbf{22.89} & \textbf{41.60} & 26.80  \\
    &  \multidds-S & \textbf{25.52} & $\textbf{12.20}^*$ & $\textbf{19.11}^*$ & $\textbf{29.37}^*$ & $\textbf{29.35}^*$ & 22.81 & 22.78 & 41.55 & \textbf{27.03}  \\
    \midrule
    \multirow{2}{*}{O2M} & Temp. & 16.61 & \textbf{6.66} & 11.29 & \textbf{21.81} & 18.60 & 11.27 & 14.92 & 32.10 & 16.26 \\
    &  \multidds-S & \textbf{17.32} & 6.59 & $\textbf{12.39}^*$ & 21.65 & $\textbf{20.61}^*$ & \textbf{11.58} & $\textbf{15.26}^*$ & $\textbf{33.52}^*$ & $\textbf{16.98}^*$ \\
    \midrule
      &   &  & \textbf{bos} & \textbf{mar} & \textbf{hin} & \textbf{mkd} & \textbf{ell} & \textbf{bul} & \textbf{fra} & \textbf{kor}  \\
    \midrule
    \multirow{2}{*}{M2O} &  Prop.   & 26.68 & 23.43 & 10.10 & 22.01 & 31.06 & $\textbf{35.62}^*$ & $\textbf{36.41}^*$ & $\textbf{37.91}^*$ & \textbf{16.91}  \\
    &  \multidds-S & \textbf{27.00} & $\textbf{25.34}^*$ & \textbf{10.57} & $\textbf{22.93}^*$ & $\textbf{32.05}^*$ & 35.27 & 35.77 & 37.30 & 16.81  \\
    \midrule
    \multirow{2}{*}{O2M} &  Temp. & 17.94 & $\textbf{14.73}^*$ & \textbf{4.93} & 15.49 & 20.59 & 24.82 & 26.60 & $\textbf{29.74}^*$ & 6.62  \\
    &  \multidds-S & \textbf{18.24} & 14.02 & 4.76 & $\textbf{15.68}^*$ & \textbf{21.44} & $\textbf{25.69}^*$ & $\textbf{27.78}^*$ & 29.60 & $\textbf{7.01}^*$  \\
    \midrule
    \end{tabular}
    \caption{BLEU scores of the best baseline and \multidds-S for all translation settings. \multidds-S performs better on more languages. For each setting, bold indicates the highest value, and $*$ means the gains are statistically significant with $p < 0.05$. }
    \label{tab:detail_bleu}
\end{table*}

\subsection{Prioritizing what to Optimize}
\label{sec:priority}


Prior works on multilingual models generally focus on improving the average performance of the model on all supported languages~\citep{massive_wild,xlmr}.
The formulation of \multidds~ reflects this objective by defining the aggregation of $n$ dev sets using \autoref{eqn:dev_agg}, which is simply the average of dev risks. However, average performance might not be the most desirable objective under all practical usage settings.
For example, it may be desirable to create a more \emph{egalitarian} system that performs well on all languages, or a more \emph{specialized} system that does particularly well on a subset of languages.

In this section, we examine the possibility of using \multidds~to control the priorities of the multilingual model by defining different dev set aggregation methods that reflect these priorities.
To do so, we first train the model for 10 epochs using regular \multidds, then switch to a different dev set aggregation method. 
Specifically, we compare \multidds~with three different priorities:

\begin{description}
\itemsep-1mm
\item[Regular:] this is the standard \multidds~that optimizes all languages throughout training using the average dev risk aggregation in \autoref{eqn:dev_agg}
\item[Low:] a more egalitarian system that optimizes the average of the four languages with the worst dev perplexity, so that \multidds~can focus on optimizing the low-performing languages
\item[High:] a more specialized system that optimizes the four languages with the best dev perplexity, for \multidds~to focus on optimizing the high-performing languages
\end{description}

\begin{table}[htbp]
    \centering
    \small
    \begin{tabular}{l|l|lll}
    \toprule
      \multirow{2}{*}{\textbf{Setting}} &  \multirow{2}{*}{\textbf{Baseline}}  & \multicolumn{3}{c}{\textbf{\multidds-S}} \\
     &  & \textbf{Regular} & \textbf{Low} & \textbf{High} \\
   \midrule 
    M2O & 26.68 & 27.00 & 26.97 & 27.08 \\
    O2M & 17.94 & 18.24 & 17.95 & 18.55 \\
    \bottomrule
    \end{tabular}
    \caption{Average BLEU of the best baseline and three \multidds-S~settings for the Diverse group. \multidds-S always outperform the baseline.}
    \label{tab:ave_bleu}
\end{table}

\begin{figure}[htbp]
  \includegraphics[width=\columnwidth]{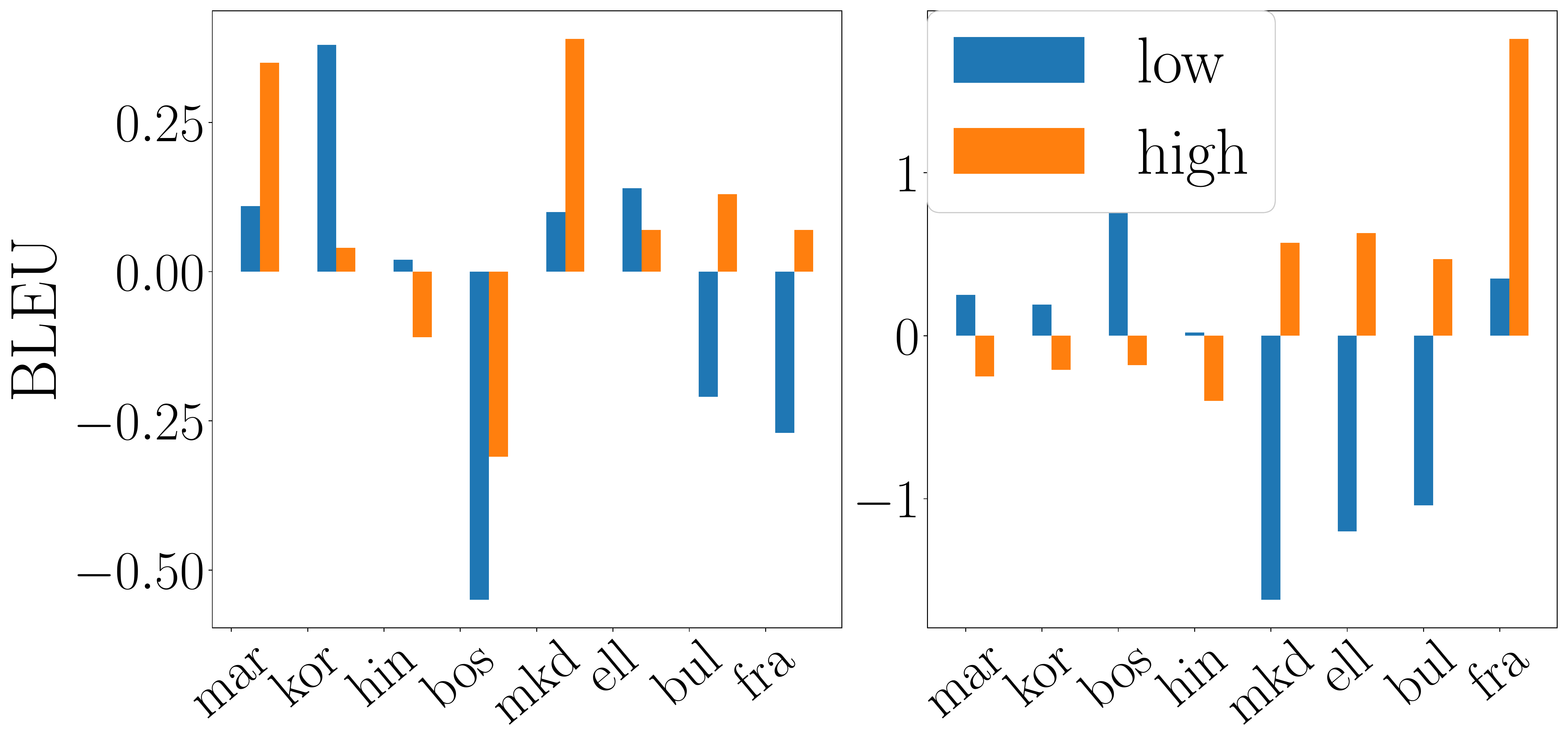}
  \captionof{figure}{\label{fig:diverse_control}The difference between Low and High optimization objectives compared to Regular for the Diverse language group. \multidds~successfully optimize for different priorities. \textit{left}: M2O; \textit{right}: O2M.}
\end{figure}

We performed experiments with these aggregation methods on the Diverse group, mainly because there is more performance trade-off among these languages.
First, in \autoref{tab:ave_bleu} we show the average BLEU over all languages, and find that \multidds~with different optimization priorities still maintains competitive average performance compared to the baseline.
More interestingly, in \autoref{fig:diverse_control}, we plot the BLEU score difference of High and Low compared to Regular for all $8$ languages. The languages are ordered on the $x$-axis from left to right in decreasing perplexity. Low generally performs better on the low-performing languages on the left, while High generally achieves the best performance on the high-performing languages on the right, with results most consistent in the O2M setting. This indicates that \multidds~is able to prioritize different predefined objectives.

It is also worth noting that low-performing languages are not always low-resource languages. For example, Korean~(\texttt{kor}) has the largest amount of training data, but its BLEU score is among the lowest. This is because it is typologically very different from English and the other training languages.
\autoref{fig:diverse_control} shows that Low is still able to focus on improving \texttt{kor}, which aligns with the predefined objective.
This fact is not considered in baseline methods that only consider data size when sampling from the training datasets.

\subsection{Learned Language Distributions}

\begin{figure}[t]
  \includegraphics[width=\columnwidth]{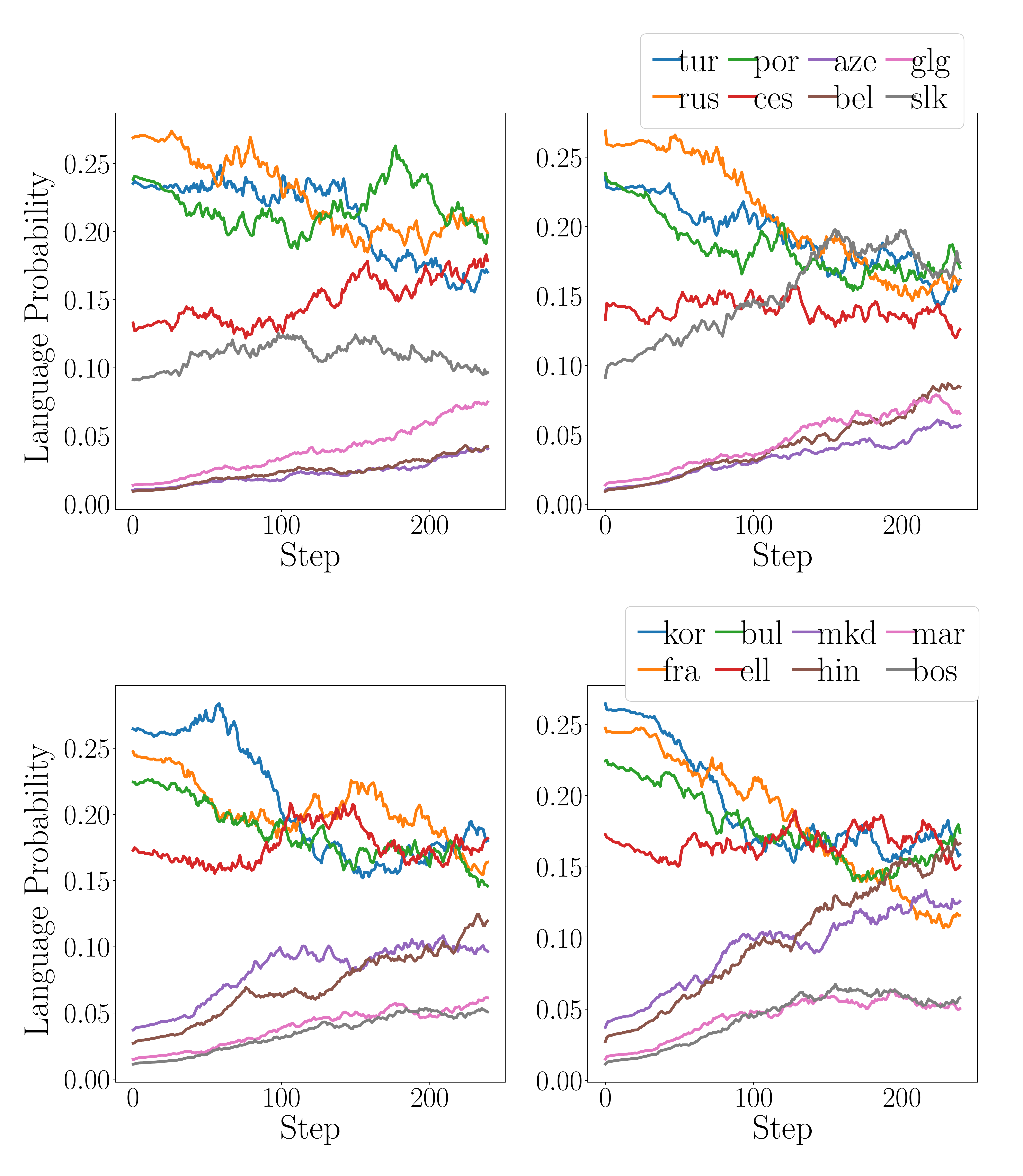}
  \captionof{figure}{\label{fig:nmt_distrib_hs}Language usage by training step. \textit{Left}: many-to-one; \textit{Right}: one-to-many; \textit{Top}: related language group; \textit{Bottom}: diverse language group.}
  \vspace{-0.1cm}
\end{figure}

In \autoref{fig:nmt_distrib_hs}, we visualize the language distribution learned by \multidds~throughout the training process. Under all settings, \multidds~gradually increases the usage of LRLs. Although initialized with the same distribution for both one-to-many and many-to-one settings, \multidds~learns to up-sample the LRLs more in the one-to-many setting, likely due to the increased importance of learning language-specific decoders in this setting. For the Diverse group, \multidds~learns to decrease the usage of Korean~(kor) the most, probably because it is very different from other languages in the group.

\subsection{\label{sec:variance}Effect of Stablized Rewards}

\begin{figure}[ht]
  \includegraphics[width=\columnwidth]{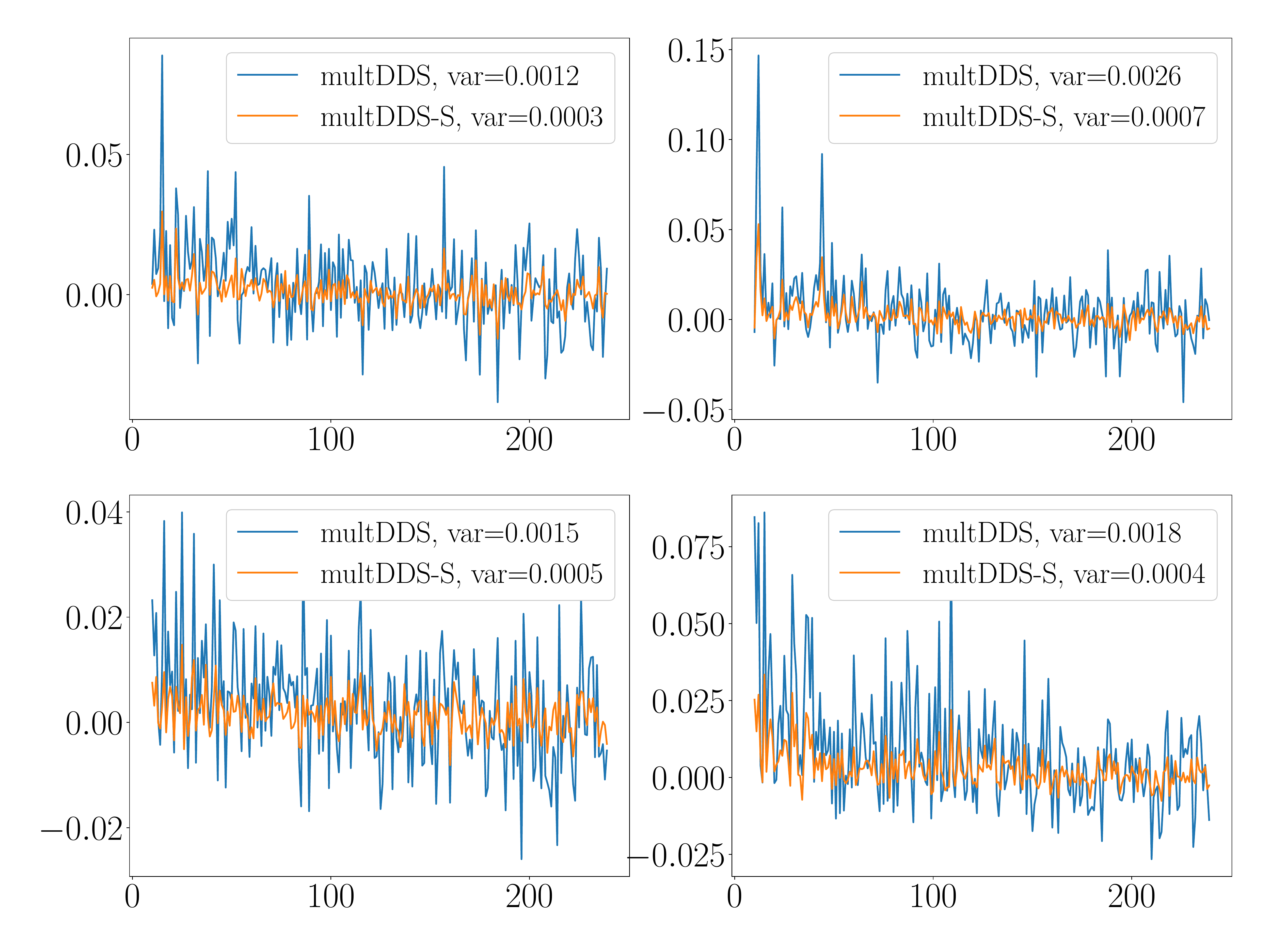}
  \captionof{figure}{\label{fig:variance}Variance of reward.  \textit{Left}: M2O; \textit{Right}: O2M; \textit{Top}: Related language group; \textit{Bottom}: Diverse language group.}
  \vspace{-0.1cm}
\end{figure}

Next, we study the effect of the stablized reward proposed in \autoref{sec:method}. In \autoref{fig:variance}, we plot the regular reward~(used by \multidds) and the stable reward~(used by \multidds-S) throughout training. For all settings, the reward in \multidds~and \multidds-S follows the similar trend, while the stable reward used in \multidds-S has consistently less variance.   

\multidds-S also results in smaller variance in the final model performance. We run \multidds~and \multidds-S with 4 different random seeds, and record the mean and variance of the average BLEU score. \autoref{tab:bleu_var} shows results for the Diverse group, which indicate that the model performance achieved using \multidds-S has lower variance and a higher mean than \multidds.

\begin{table}[t]
    \centering
    \small
    \begin{tabular}{l|ll|ll}
    \toprule
    \multirow{2}{*}{\textbf{Method}} & \multicolumn{2}{c}{\textbf{M2O}} &  \multicolumn{2}{c}{\textbf{O2M}} \\
     & \textbf{Mean} & \textbf{Var.} & \textbf{Mean} & \textbf{Var.} \\
    \midrule
     \multidds   & 26.85  & 0.04 & 18.20 & 0.05 \\
     \multidds-S  & 26.94  & 0.02 & 18.24 & 0.02 \\
    \bottomrule
    \end{tabular}
    \caption{Mean and variance of the average BLEU score for the Diverse group. The models trained with \multidds-S perform better and have less variance.}
    \label{tab:bleu_var}
\end{table}

\begin{figure}[t]
  \includegraphics[width=\columnwidth]{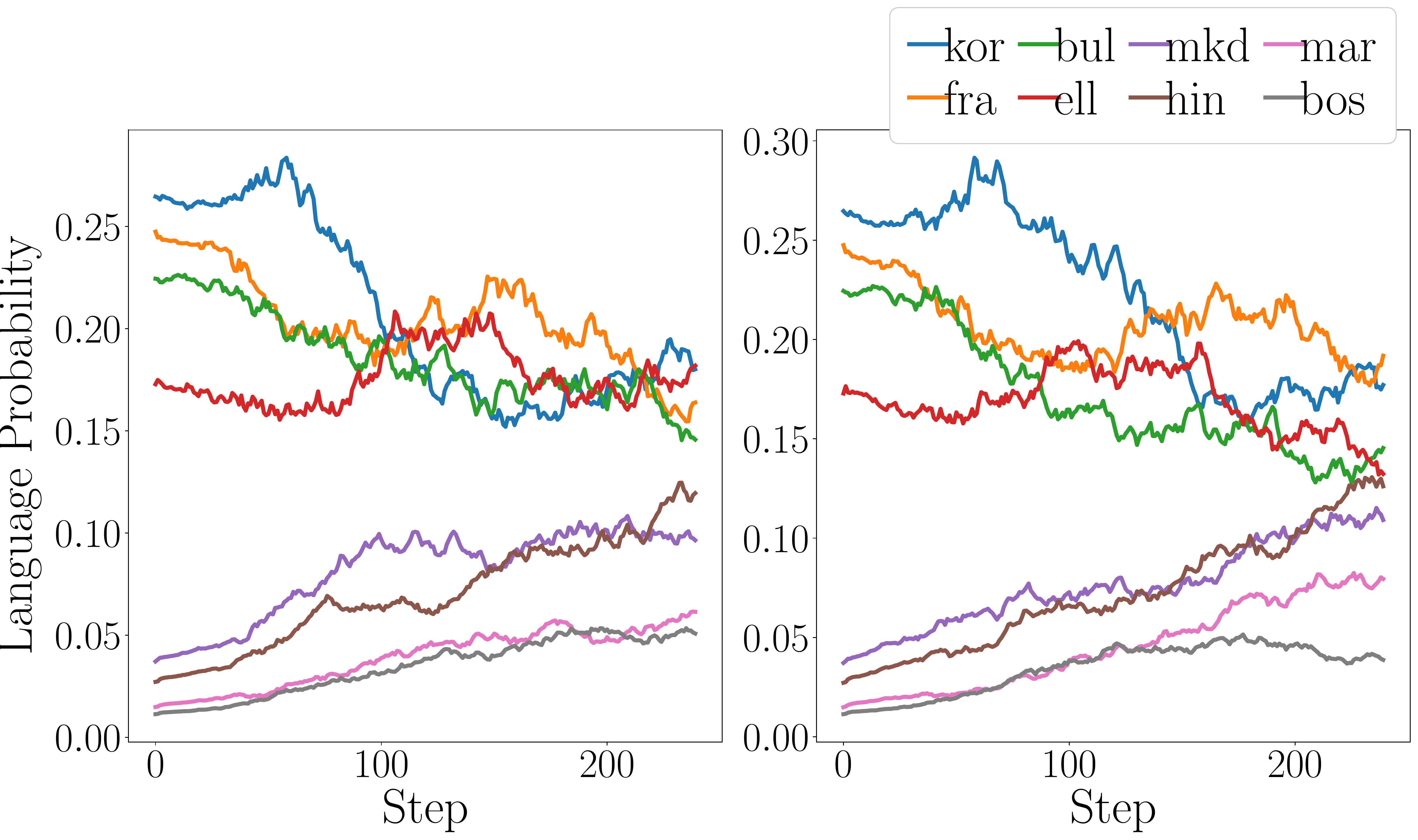}
  \captionof{figure}{\label{fig:reg_vs_stable}Language usage for the M2O-Diverse setting. \textit{Left}: \multidds-S; \textit{Right}: \multidds. The two figures follow similar trends while \multidds~changes more drastically.}
  \vspace{-0.1cm}
\end{figure}

 Additionally, we compare the learned language distribution of \multidds-S and \multidds~in \autoref{fig:reg_vs_stable}. The learned language distribution in both plots fluctuates similarly, but \multidds~has more drastic changes than \multidds-S. This is also likely due to the reward of \multidds-S having less variance than that of \multidds. 

\section{\label{related} Related Work}
Our work is related to the multilingual training methods in general.
Multilingual training has a rich history \cite{schultz1998multilingual,polylingual_topic_model,crosslingual_classify,tackstrom2013token}, but has become particularly prominent in recent years due the ability of neural networks to easily perform multi-task learning \cite{dong-etal-2015-multi,plank-etal-2016-multilingual,johnson16multilingual}.
As stated previously, recent results have demonstrated the importance of balancing HRLs and LRLs during multilingual training \citep{massive_wild,xlmr}, which is largely done with heuristic sampling using a temperature term; \multidds~provides a more effective and less heuristic method.
\citet{tcs,choose_transfer} choose languages from multilingual data to improve the performance on a particular language, while our work instead aims to train a single model that handles translation between many languages. \citep{adaptive_routing,three_one_to_many,sde} propose improvements to the model architecture to improve multilingual performance, while \multidds~is a model-agnostic and optimizes multilingual data usage.

Our work is also related to machine learning methods that balance multitask learning~\citep{gradnorm,multitask_loss_weight}.
For example, \citet{multitask_loss_weight} proposes to weigh the training loss from a multitask model based on the uncertainty of each task. Our method focuses on optimizing the multilingual data usage, and is both somewhat orthogonal to and less heuristic than such loss weighting methods. Finally, our work is related to meta-learning, which is used in hyperparameter optimization~\citep{hyper_grad}, model initialization for fast adaptation~\citep{finn2017model}, and data weighting~\citep{learn_reweight}.
Notably, \citet{gu-etal-2018-meta} apply meta-learning to learn an NMT model initialization for a set of languages, so that it can be quickly fine-tuned for any language.
This is different in motivation from our method because it requires an adapted model for each of the language, while our method aims to optimize a single model to support all languages.
To our knowledge, our work is the first to apply meta-learning to optimize data usage for multilingual objectives.  
\section{Conclusion}
In this paper, we propose \multidds, an algorithm that learns a language scorer to optimize multilingual data usage to achieve good performance on many different languages. We extend and improve over previous work on DDS~\citep{dds}, with a more efficient algorithmic instantiation tailored for the multilingual training problem and a stable reward to optimize multiple objectives. \multidds~not only outperforms prior methods in terms of overall performance on all languages, but also provides a flexible framework to prioritize different multilingual objectives.

Notably, \multidds~is not limited to NMT, and future work may consider applications to other multilingual tasks.
In addition, there are other conceivable multilingual optimization objectives than those we explored in \autoref{sec:priority}.  

\section{Acknowledgement}
The first author is supported by a research grant from the Tang Family Foundation. This work was supported in part by NSF grant IIS-1812327. The authors would like to thank Amazon for providing GPU credits.

\bibliography{main}
\bibliographystyle{acl_natbib}

\clearpage
\newpage
\appendix
\section{Appendix}

\subsection{\label{app:stepahead}Effect of Step-ahead Reward}

\begin{table}[ht]
    \centering
    \resizebox{.4\textwidth}{!}{
    \begin{tabular}{l|l|ll}
    \toprule
     \multirow{2}{*}{Setting} &  \multirow{2}{*}{Baseline}  & \multicolumn{2}{c}{\multidds} \\
    & & Moving Ave. & Step-ahead \\
   \midrule 
     M2O & 24.88 & 25.19 & 25.26  \\
     O2M & 16.61 & 17.17 & 17.17 \\
    \bottomrule
    \end{tabular}
    }
    \caption{Ave. BLEU for the Related language group. The step-ahead reward proposed in the paper is better or comparable with the moving average, and both are better than the baseline.}
    \label{tab:stepahead_ave_bleu}
\end{table}

\subsection{\label{app:hparams}Hyperparameters}
In this section, we list the details of preprocessing and hyperparameters we use for the experiments.
\begin{itemize}
    \item We use 6 encoder and decoder layers, with 4 attention heads
    \item The embedding size is set to 512, and the feed-forward layer has a dimension of 1024
    \item We use the dropout rate of 0.3
    \item The batch size is set to 9600 tokens
    \item We use label smoothing with rate of 0.1
    \item We use the scaled $l_2$ normalization before residual connection, which is shown to be helpful for small data~\citep{transformer_no_tears}
\end{itemize}

\subsection{\label{app:stats}Dataset statistics}

\begin{table}[ht]
  \centering
  \begin{tabular}{c|ccc}
  \toprule
  \textbf{Language} & \textbf{Train} & \textbf{Dev} & \textbf{Test}  \\
  \midrule
  aze & 5.94k &  671 &  903  \\
  bel & 4.51k &  248 &  664  \\
  glg & 10.0k &  682 & 1007  \\
  slk & 61.5k & 2271 & 2445  \\
  tur & 182k  & 4045 & 5029  \\
  rus & 208k  & 4814 & 5483 \\
  por & 185k  & 4035 & 4855 \\
  ces & 103k  & 3462 & 3831 \\
  \bottomrule
  \end{tabular}
  \vspace{0.2cm}
  \caption{\label{tab:related_data}Statistics of the related language group.}
\end{table} 

\begin{table}[ht]
  \centering
  \begin{tabular}{c|ccc}
  \toprule
  \textbf{Language} & \textbf{Train} & \textbf{Dev} & \textbf{Test}  \\
  \midrule
  bos & 5.64k &  474 &  463  \\
  mar & 9.84k &  767 &  1090  \\
  hin & 18.79k &  854 & 1243  \\
  mkd & 25.33k & 640 & 438  \\
  ell & 134k  & 3344 & 4433  \\
  bul & 174k  & 4082 & 5060 \\
  fra & 192k  & 4320 & 4866 \\
  kor & 205k  & 4441 & 5637 \\
  \bottomrule
  \end{tabular}
  \vspace{0.2cm}
  \caption{\label{tab:diverse_data}Statistics of the diverse language group.}
\end{table}

\subsection{Detailed Results for All Settings}

\begin{table*}[ht]
    \centering
    \begin{tabular}{l|l|cccccccc}
    \toprule
       Method  & Avg. & aze & bel & glg & slk & tur & rus & por & ces  \\
    \midrule
      Uni. ($\tau=\mathcal{1}$) & 22.63 & 8.81 & 14.80 & 25.22 & 27.32 & 20.16 & 20.95 & 38.69 & 25.11  \\
      Temp. ($\tau=5$)  & 24.00 & 10.42 & 15.85 & 27.63 & 28.38 & 21.53 & 21.82 & 40.18 & 26.26   \\
      Prop. ($\tau=1$)   & 24.88 & 11.20 & 17.17 & 27.51 & 28.85 & 23.09 & 22.89 & 41.60 & 26.80  \\
    \midrule
      \multidds   & 25.26 & 12.20 & 18.60 & 28.83 & 29.21 & 22.24 & 22.50 & 41.40 & 27.22 \\
      \multidds-S & \textbf{25.52} & 12.20 & 19.11 & 29.37 & 29.35 & 22.81 & 22.78 & 41.55 & 27.03  \\
    \bottomrule
    \end{tabular}
    \caption{BLEU score of the baselines and our method on the Related language group for many-to-one translation}
    \label{tab:related_m2o}
\end{table*}

\begin{table*}[ht]
    \centering
    \begin{tabular}{l|l|cccccccc}
    \toprule
       Method  & Avg. & bos & mar & hin & mkd & ell & bul & fra & kor  \\
    \midrule
      Uni. ($\tau=\mathcal{1}$) & 24.81 & 21.52 & 9.48 & 19.99 & 30.46 & 33.22 & 33.70 & 35.15 & 15.03   \\
      Temp. ($\tau=5$)  &  26.01 & 23.47 & 10.19 & 21.26 & 31.13 & 34.69 & 34.94 & 36.44 & 16.00 \\
      Prop. ($\tau=1$)  & 26.68 & 23.43 & 10.10 & 22.01 & 31.06 & 
      35.62 & 36.41 & 37.91 & 16.91  \\
    \midrule
      \multidds   & 26.65 & 25.00 & 10.79 & 22.40 & 31.62 & 34.80 & 35.22 & 37.02 & 16.36 \\
      \multidds-S & \textbf{27.00} & 25.34 & 10.57 & 22.93 & 32.05 & 35.27 & 35.77 & 37.30 & 16.81  \\
    \bottomrule
    \end{tabular}
    \caption{BLEU score of the baselines and our method on the Diverse language group for many-to-one translation}
    \label{tab:diverse_m2o}
\end{table*}

\begin{table*}[ht]
    \centering
    \begin{tabular}{l|l|cccccccc}
    \toprule
       Method  & Avg. & aze & bel & glg & slk & tur & rus & por & ces  \\
    \midrule
     Uni. ($\tau=\mathcal{1}$) & 15.54 & 5.76 & 10.51 & 21.08 & 17.83 & 9.94 & 13.59 & 30.33 & 15.35 \\
     Temp. ($\tau=5$) & 16.61 & 6.66 & 11.29 & 21.81 & 18.60 & 11.27 & 14.92 & 32.10 & 16.26 \\
     Prop. ($\tau=1$) & 15.49 & 4.42 & 5.99 & 14.92 & 17.37 & 12.86 & 16.98 & 34.90 & 16.53 \\
    \midrule
      \multidds & 17.17 & 6.24 & 11.75 & 21.46 & 20.67 & 11.51 & 15.42 & 33.41 & 16.94 \\
      \multidds-S & \textbf{17.32} & 6.59 & 12.39 & 21.65 & 20.61 & 11.58 & 15.26 & 33.52 & 16.98 \\
    \bottomrule
    \end{tabular}
    \caption{BLEU score of the baselines and our method on the Related language group for one-to-many translation}
    \label{tab:related_o2m}
\end{table*}

\begin{table*}[ht]
    \centering
    \begin{tabular}{l|l|cccccccc}
    \toprule
       Method  & Avg. & bos & mar & hin & mkd & ell & bul & fra & kor  \\
    \midrule
      Uni. ($\tau=\mathcal{1}$) & 16.86 & 14.12 & 4.69 & 14.52 & 20.10 & 22.87 & 25.02 & 27.64 & 5.95  \\
      Temp. ($\tau=5$) & 17.94 & 14.73 & 4.93 & 15.49 & 20.59 & 24.82 & 26.60 & 29.74 & 6.62  \\
      Prop. ($\tau=1$) & 16.79 & 6.93 & 3.69 & 10.70 & 15.77 & 26.69 & 29.59 & 33.51 & 7.49 \\
    \midrule
      \multidds & \textbf{18.40} & 14.91 & 4.83 & 14.96 & 22.25 & 24.80 & 27.99 & 30.77 & 6.75 \\
      \multidds-S & 18.24 & 14.02 & 4.76 & 15.68 & 21.44 & 25.69 & 27.78 & 29.60 & 7.01  \\
    \bottomrule
    \end{tabular}
    \caption{BLEU score of the baselines and our method on the Diverse language group for one-to-many translation}
    \label{tab:diverse_o2m}
\end{table*}

\end{document}